# Are Transformers Truly Foundational for Robotics?

James A. R. Marshall[1,2,*] and Andrew B. Barron[3]

[1]Centre for Machine Intelligence and Department of Computer Science, University of Sheffield, S10 2TN, UNITED KINGDOM
[2]Opteran Technologies Ltd., The Innovation Centre, 217 Portobello, S1 4DP, UNITED KINGDOM
[3]School of Natural Sciences, Macquarie University, 14 Eastern Road, 2019 NSW, AUSTRALIA
[*]Address for correspondence: james.marshall@sheffield.ac.uk

# Abstract

Generative Pre-Trained Transformers (GPTs) are hyped to revolutionize robotics. Here we question their utility. GPTs for autonomous robotics demand enormous and costly compute, excessive training times and (often) offboard wireless control. We contrast GPT state of the art with how tiny insect brains have achieved robust autonomy with none of these constraints. We highlight lessons that can be learned from biology to enhance the utility of GPTs in robotics.

# Introduction

Recent years have seen major advances in generative Artificial Intelligence due to the development and deployment of a new architecture; the Generative Pre-Trained Transformer (GPT), or transformer for short [1]. Through adding an attentional mechanism to deep neural networks and deploying on internet-scale training sets, transformers have led to rapid advancements in Large Language Models (LLMs) for natural language processing and generation. Following these early applications, transformers have, alongside other architectures such as diffusion models [2] been applied to the development of Visual Language Models for text to image and video (e.g. [3]) as well as other multimodal applications (e.g. [4]). These successes have inspired the investigation of transformer architectures for the robotics domain. The challenges of unstructured multimodal inputs sensed in complicated environments, coupled with high degrees of freedom in robot control, have constrained the development of robots that are simultaneously generally capable, and robust, in their behaviour. The promise of transformers for robotics appears to be that large-scale training can, through specialisation on further smaller-scale training sets, provide general and adaptable solutions to a wide variety of robotics tasks [5]. Because they can be applied across so many application domains transformer-based approaches have been labelled Foundation Models [5] indicating their supposed fundamental status but also their incomplete nature. Applications of foundation models to robotics have recently taken off in the minds of developers and researchers.

Transformers have their genesis in large language modelling (LLM). LLMs have also proved to be generalizable and transformative to many applications, but they are not without limitations. As we review below, there are increasingly recognised issues with LLMs in the areas of training dataset size, compute resources for training, the financial and ecological costs of both, as well as robustness of behavioural output. In this article we question whether transformer architectures are likely to be truly foundational for robotics. We ask whether transformers provide the only or best route towards Artificial General Autonomy, proposing that, unlike 'intelligence', [6] the level of autonomy of a robotics system is well-defined, measurable, and economically meaningful.

Drawing on earlier critiques of GPTs and related approaches, we argue that transformers provide a facsimile of autonomy rather than true autonomy. We then review alternative approaches that have been proposed. The contrast between GPT solutions to autonomous robotics and biological solutions to autonomous behavioural control achieved by animal brains is stark. We explore this contrast to propose what is missing from current GPT approaches, and what could be added in to enhance robust and scalable robot autonomy.

# Progress in Applying Transformer Architectures to Autonomy

Transformers have seen rapid application to robot autonomy. As well as high profile commercial announcements and demonstrations, end-to-end solutions to robot autonomy have been developed in the peer-reviewed literature by both academic and industrial groups, to tasks particularly focussing on robot navigation and dexterity (for a review, see [7]).

While the early promise of transformers for robot autonomy seems to be being realised, for a general and scalable solution it is essential to recognise that this technology still comes with significant limitations that will constrain future performance and adoption. While these are active areas of research, and some of these may become less acute as the traditional efficiencies associated with the development and deployment of a novel technology are realised, we argue that there are fundamental structural issues with current transformer architectures, and that these should motivate a longer term search for alternative and complementary approaches, which we review later in this article.

## Training Data Size and Cost Requirements are Likely to Grow

At the heart of the transformer approach to any problem is a scaling requirement. Given the lack of inductive biases these learning systems are highly flexible, however the corollary of this is that their training data requirements are vast. The usual approach for deployment of a transformer-based foundation model is to train on an internet-scale corpus so that the model acquires multi-modal correspondences and domain knowledge, then further specialise on a smaller training data set for a specific set of tasks. The costs of this are very substantial. Even excluding environmental impacts, state-of-the-art LLMs cost on the order of $10s to $100s of millions per training episode [8], although rapid reductions in training and inference costs are being made [9]. For robotics applications, further training for particular tasks such as navigation and manipulation is usually required. The availability and cost of acquiring good training datasets is recognised as a major problem. Proposed solutions include the curation of open datasets covering multiple tasks and robot types [10], although currently these can be biased to a relatively small number of tasks. There is also the extensive use of physics-based simulators to generate training data (e.g. [11]). We argue that, similarly to LLMs, exponentially increasing quantities of data are likely to be required to sustain advances in performance [12]. Even for text and multimodal datasets where the internet provides a very large corpus of training data 'for free', the availability of training data risks becoming a limiting factor [13]. For robotics datasets the costs of collecting useful training data, either physically or through simulation, will be much more acute; replacing physical data collection with collection via simulation simply trades one kind of resource – experimental time – with another – computational time, albeit with the latter being more scalable. Furthermore, since improvement in transformers' performance is predicated on increases in scale of training data and weights this problem will only get worse – for example, in multimodal AI 'zero-shot' generalization has been shown to have exponential training data requirements [14].

## Compute and Infrastructure Costs and Requirements will Persist

Once the costs of training a transformer-based architecture are paid, the inference costs at deployment can still be substantial. For example, Meta's Llama 3.1 has cloud-scale deployments (405bn double-precision parameters). There are reduced size and precision versions suitable for deployment on local GPUs (e.g. 8bn half-precision integer parameters), which can take ~20-100GB of memory for inference. This demands a substantial GPU for even the simplest models being run on a robot [15]. While binarisation, quantisation, and other approaches have been used to help design edge AI accelerators for deep and convolutional neural networks (e.g. [16]), the scale of the problem for transformers is many orders of magnitude larger. For example, for one of the longest researched applications of deep nets, object detection, one state of the art algorithm has on the order of 10m-80m network weights [15], compared to the 8bn-405bn weights mentioned above for a state-of-the-art LLM. This represents a four orders of magnitude difference in scale, even before the additional requirements of training a transformer for robotics tasks are taken into account. Hence there is very active research into methods to avoid the cloud compute bottleneck, including utilisation of novel technologies such as 6G, [17]. Moore's law and the advent of novel parallel compute architectures has traditionally saved AI, and computer software more generally. For foundation models, however, we argue that although available compute can be scaled exponentially, the exponential requirements for model size and throughput will be in opposition. A real-terms reduction in requirements for compute as performance improvements are sought will only occur when the exponent for the former is greater than the exponent for the latter. However growing evidence suggests we are moving to a post-Moore's Law world where further innovation in materials is required to make progress [e.g. [18]]. Even if compute could still scale faster than data requirements scale, given the potential for ongoing algorithmic improvements [9], below we argue that there are very many orders of magnitude difference between the training that is achievable for a transformer, and the 'training' that has genuinely solved autonomy over evolutionary time.

### Hallucinations for Transformers in Robotics May Become Acute

As a consequence of their statistical training and inference, LLMs are prone to confabulation and hallucination, defined as producing outputs that are inconsistent with user input and/or world knowledge and common sense [19,20]. While such outputs can still be damaging even for a disembodied AI, for example in the social and political arenas [21] when a transformer architecture is embodied the risks are magnified, much as acting on hallucinatory perceptions and impulses in human mental illness can lead to recognized harms to self and others. As with humans, hallucinations may manifest in ways likely to cause harm to the robot or to others, and adversarial attacks on guardrails for transformers in robotics have already been demonstrated [22]. While mitigation of hallucinations is an ongoing area of research [20], as others also argue [23] we contend that the fundamentally correlational nature of transformers will render hallucinations inescapable. Failures of reasoning are also inherent in symbolic reasoning by GPTs [24] and chain of reasoning models [25]. This is likely to require that humans remain in the control loop as teleoperators to ensure robots are remotely supervised, or that robots are isolated from humans, or both. Any of these outcomes will of course limit the promised benefits of robotics. As other researchers have argued, these structural issues with statistical approaches to AI are unlikely to find remedy without significant architectural change [26].

## Transformers Give a Facsimile of Intelligent Autonomy

Given the above concerns, why are transformers seeing increasing adoption for robotics? We attribute this to two factors: first, as with LLMs and VLMs, striking early advances have been made in traditionally very difficult areas, such as humanoid control, manipulation, and, of course, natural language interfaces. Second, however, we believe a tendency of human observers to anthropomorphise often leads some of them to ascribe abilities, and the potential for understanding, that the architecture does not, and cannot, technically support.

While there are many types of transformer the central motif is a repeating unit composed of a self-attention block followed by a multilayer perceptron block [27] (Figure 1, right). The control flow is feedforward, while the attention mechanism learns which earlier elements of the input to attend to in predicting the next appropriate action. As with LLMs, both the power and generalizability of transformers for robotics comes from their extensive training so that, once trained, they can perform the operation of matching an input to a predicted output. In robotics transformers succeed in resolving and executing an action from an input, but this is achieved by interpolation and extrapolation of the training set, with unreliable off-training-set performance [28]. There is no reasoning and no reason why a transformer selects one response over another, other than the selected option carrying the highest predictive weight following training [29]. The same can be said of the language abilities of LLMs, which have been described as stochastic parrots [30].

Training and reference to learned experience is an important part of biological autonomous decision making too, but for humans and other animals decision making is also supported by reasoning from models of how the world works, how other involved agents should operate, and why the selected action is situation appropriate [31]. Transformers lack these models [24,32]. An autonomous robot's capacity will be limited by the scope of the training dataset. Since transformers responses are unreasoned products of the training data, any transformer-based application cannot justify a decision other than by statistical association to the training data. This poses serious challenges for any form of human / robot interaction. If we were to ask a well-intentioned human coworker why they made an error they would do their best to explain the reasoning behind their actions [33]. If we ask a transformer based robot why it made an error there would be no reasoned answer *per se*; the answer to the query will have at best a correlation but no causal relationship to the error made, and subject to hallucination as described above. Reasoning from models of how the world works can allow forms of introspection and metacognition that can interrogate why a wrong choice has been made, or query wrong decisions before any action is taken. We contend that feedforward transformer-based applications are structurally incapable of reliable metacognition [29,31].

## Alternatives and Complements to Transformers for Autonomy

If transformers are not the full answer, what is? Here we review the main alternative proposals, with an emphasis on our preferred approach, drawing deep inspiration from how the biological brain solves the autonomy problem.

### Natural Intelligence

The gulf between transformer approaches to robotics and how biological brains produce autonomous behaviour is stark (Figure 1). Most often comparisons are drawn between LLMs, GPTs and human reasoning [29,31,34], but the comparison with animal brains and animal reasoning is even more pronounced. For example, the honey bee brain is tiny (just over one cubic millimeter) and contains fewer than one million neurons [35]. The number of synapses in the bee brain is not known, but if we can infer from the *Drosophila* connectome [36] there will be fewer than half a billion synapses in the bee brain. (Figure 1, left). Demonstrably, this is all a bee needs to reliably navigate over long (several kilometre) distances, autonomously harvest pollen and nectar from the environment, communicate and coordinate their efforts with their hive mates, and perform all the many jobs needed to build and maintain their colony, including raising the next generation. They can solve complex foraging economics problems, majoring on the resources their colony needs and harvesting them from cryptic and ephemeral flowers patchily distributed in the environment [37]. Bees are able to fly with no practice, and just twenty minutes of structured flight time around the hive is enough for them to be able to navigate proficiently in their environment [38]. The contrast with the prolonged training needed by transformers could not be greater. The power consumption of a bee brain as it performs entirely on-board autonomous decision making is infinitesimal compared to any GPT. In contrast to transformers, animal brains have been massively 'pre-trained' on a planetary scale, to use minimal information and generate a very wide variety of behaviours (Figure 2). It is trivial to observe that the scale of this evolutionary pre-training, spanning very many trillions of instantiations of tens of millions of different species, across hundreds of millions of years, cannot be matched by computational approaches; even if it could, arguably we do not have a sufficiently robust and evolvable representation to match the genetic language that encodes for body and brain morphology and behaviour in nature [39]. But the greater point is this: we don't need to match the *process* by which bee intelligence evolved if we want to match the performance of that evolved intelligence. That can be done by studying just the end point of the evolutionary process – the embodied bee brain.

How, then, does the humble bee outperform transformers in compute, energetic cost, and training time? In a word – structure. The generalisability of transformers, and arguably their elegance, is because before pre-training they are not structurally differentiated according to function. The insect brain, by contrast is a case study in structure-function specialization. The insect brain is subdivided into modules (Figure 1, left). Each module is specialised for processing different domains of the autonomous decision-making challenge. Each specialization in each module exploits the regularities and properties of the information it is processing to reduce compute and increase overall system efficiency. For example, specialized modules in the bee, ant and fly brain process the pattern of polarized light in the sky generated around the sun [40,41]. This is a valuable and robust navigational cue. Its structure is preserved by a topographic processor – the protocerebral bridge in the central complex – which outputs to a region that operates as a ring attractor to establish orientation of the animal relative to external cues [40,42,43]. This connects to yet another module which is topographically structured as the azimuth, and can support the relative localization of the insect to external objects [40,41]. The regularities of the external world are reflected in how they are represented in the insect brain, which conveys a form of intuitive physics (albeit very different from the type of physics engines used in AI). Olfactory and visual sensory lobes are each specialised to the input properties of their sensory domain. The sensory lobes sharpen, enhance and ultimately compress sensory signals for projection to multimodal sensory integration regions [44]. The largest of these, the mushroom body, has a structure similar to a three-layer neural network with an expanded middle layer [45-47]. This seems especially adept at multimodal classification.

Insects lack the declarative reasoning of humans, but their reasoning is built around a form of elementary world model. Insects possess a unitary and coherent representation of external space within which they have a first-person perspective on objects around them [48]. The valence of objects is influenced by the insect's learned experience with them, as well as innate valence and subjective physiological state [44]. Differences in valence and location of objects arbitrate the insect's selection [49-51]. This form of reasoning might be elementary, but it is still more comprehensible and explicit than the reasonless transformers. It is increasingly recognized that AI stands to benefit tremendously from importing concepts and algorithms from insect neuroscience [52,53].

## Objective AI and World Models

Other researchers have proposed that indeed the autonomy abilities of animals (including those 'simpler than humans) should provide inspiration for AI researchers [54]. However, this inspiration is much looser than the Natural Intelligence approach above. While the 'objective AI approach does indeed propose modular AI architectures that correspond with an understanding of the human brain developed in neuroscience, cognitive science, and psychology, the proposal is actually quite different; rather than directly seek to reverse-engineer neural circuits in specialist brain modules, instead the idea is to design trainable modules that interface with each other in order to generate more adaptive behaviour than a largely undifferentiated large neural net could be expected to. Thus, for example, rather than directly seek to understand how feature detectors in the early primate visual system function, a feature detector module would be trained. A key part of the proposal is the reintroduction of explicit and configurable worlds models, drawing inspiration from cognitive science; however these also remain trained from data [55].

## Hybrid Approaches

Still other researchers, drawing on a long running proposal but also gaining renewed motivation from contemporary developments in AI, have proposed the 'neurosymbolic approach' [26]. This approach argues that, while deep nets are very suitable for perceptual tasks such as object detection, they are fundamentally unsuited to the symbolic manipulation that is part of reasoning, planning, and decision making. In the context of transformers, this has recently been vindicated by observations that LLMs fail to robustly deal with and manipulate symbolic knowledge [24,32]. Thus the proposal is to combine the perceptual strengths of statistical AI with the causal strengths of the older, symbolic, approach to AI. Given the neural bases of symbolic reasoning in the brain are poorly understood, this is a particularly pragmatic approach. In doing so it is hoped that the limitations of the first, symbolic, wave of AI, will be ameliorated by working around the problems they suffered in having sole responsibility for dealing with the perceptual complexity of the real world [56]. We suggest that an even more powerful combination could include the use of Natural Intelligence approaches to perception and modelling of space and decision option sets within it.

Since transformers may inform our understanding of aspects of 'higher' cognitive function in real brains (see 'Natural Intelligence' above), they could still have great value informing a component of full autonomous system stacks grounded on basal mechanisms derived from natural brains, rather than their foundations themselves. Just as activity and learned filters in deep neural networks show interesting corollaries with function of natural brains [56], transformers do appear to capture some interesting fundamental aspects of language and visual recapitulation [57] (but see [58] and [59]), and hence may form a component of a fully autonomous system grounded on firmer foundations (see 'Hybrid Approaches' below).

# Conclusion

Transformer architectures have brought to robotics the rapid progress that they had already brought to natural language and muli-modal AI. However, there are reasons to continue the search for solutions to the robotics autonomy problem. Transformer architectures treat the world in purely statistical terms, albeit grounded in perceptual inputs. This was arguably a deliberate choice in response to the 'bitter lesson' [60], that inductive biases in AI have historically failed [56]. However, this results in an autonomy solution very different to the way the only truly autonomous artefact known to humanity, the biological brain, functions. Here we have highlighted this, and conclude by arguing that the tremendous recent advances in data on, and understanding of, a variety of brains, means the time is ripe to revisit the 'bitter lesson', and see what new lessons for AI can be learned from their study.

## Acknowledgements

We thank Sarah Moth-Lund Christensen, the editor and two anonymous reviewers for comment and feedback on the manuscript. This manuscript was partially written when the authors were participants in the Mathematics of Intelligences Long Programme, and Workshop III on Naturalistic Approaches to Intelligence, at the Institute for Pure and Applied Mathematics, UCLA. The authors are grateful for the partial financial support they received from the NSF via the programme to enable their participation. Marshall is supported by the Centre for Machine Intelligence at the University of Sheffield. Barron is supported by funding from the Templeton World Charity Foundation (TWCF-2020-0539) the Australian Research Council (DP230100006, DP240100400) and the Macquarie University Bioinnovation Initiative.

## Author Contributions

JARM and ABB conceived of, wrote, and edited the paper.

## Competing Interests

JARM is Founder Science Officer at and shareholder in Opteran Technologies Ltd. ABB is an unremunerated Academic Advisory Board member at Opteran Technologies Ltd.

## Figure Legends

Figure 1. Left: The brain of a honey bee forager *Apis mellifera* provides high levels of autonomy integrating multi-modal sensory data to navigate, communicate locations in space, learn associations between stimuli and rewards, using fewer than 960,000 neurons. Distinct brain regions specialise in perception including vision (optic lopes, yellow and orange), olfaction (antennal lobes, blue), and feed into multimodal memory centres (mushroom bodies, red). Sensory and memory pathways converge in the central complex which integrates sensing and learned associations in a single representation of the bee situated relative to percepts weighted by the bees internal state. This is sufficient to resolve competing goals, which drives behaviour directly by interfacing with premotor neurons (not shown). Brain regions are highly differentiated in structure and function according to task demands, and come together in a modular architecture with high degrees of intra-module connectivity but limited and well defined inter-module connections. Image source: insectbraindb.org Right: the generative pre-trained transformer (GPT) architecture. Multimodal sensory inputs are embedded in high dimensional space (not shown) then feed into repeated blocks of attentional mechanisms (yellow) followed by feed forward deep networks (blue), with intermediate normalisation and selection layers (not shown). Each block is hence a very large and non-sparse matrix, with matrix multiplications propagating through the GPT to produce the next output in sequence. Knowledge of the task is encoded in the learned values within the matrices, whose total entry counts typically range in the billions to trillions. Thus although total GPT parameters vastly exceed the number of synaptic connections in a simple brain, they are far less robust in behavioural output. Figure adapted from [1].

Figure 2. Transformer-based approaches to autonomy rely on internet-scale datasets as input, and are trained to process input from a suite of high resolution sensors, such as 4k cameras and LiDar, in order to provide a limited behavioural repertoire, in comparison to biological autonomous agents. In contrast, 600m years of evolution on a planetary scale, with complex physics, has encoded blueprints to build autonomous brains into the genome of a massive variety of animal species. These brains process much sparser input from specialized sensor suites, captured using

active perception and behaviour, to generate a hugely rich variety of adaptive behaviours. Image sources: NASA (Earth), insectbraindb.org (honeybee brain).

Figure 1

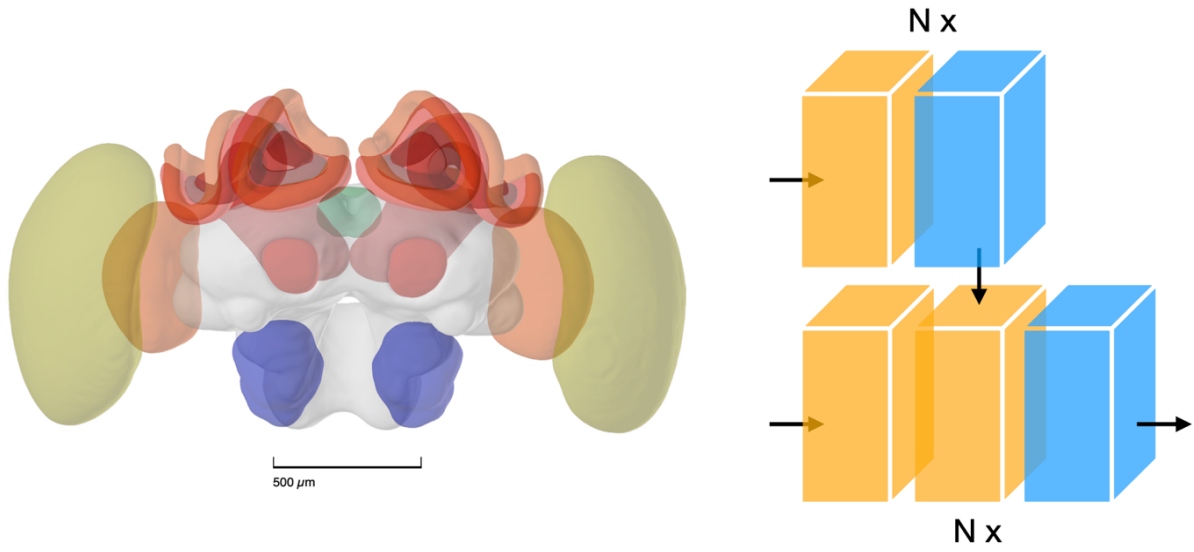

Figure 2

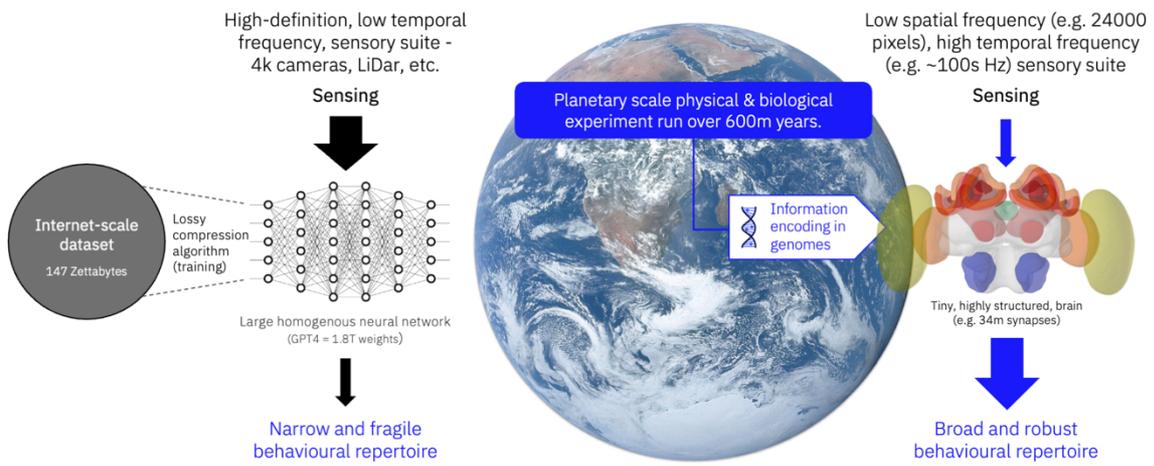